# Phishing Detection by determining Reliability Factor using Rough Set Theory.

Anugrah Kumar[1], Sanjiban Shekar Roy, Sanklan Saxena

*School Of Computing Science and Engineering*

*VIT University*

*Vellore, India*

E-Mail:[1]anugrah18@gmail.com

Sarvesh SS Rawat[2]

*School Of Electronics and Instrumentation*

*VIT University*

*Vellore, India*

E-Mail:[2]sss.sarvesh888@gmail.com

*Abstract*— **Phishing is a common online weapon, used against users, by Phishers for acquiring a confidential information through deception. Since the inception of internet , nearly everything , ranging from money transaction to sharing information, is done online in most parts of the world. This has also given rise to malicious activities such as Phishing. Detecting Phishing is an intricate process due to complexity, ambiguity and copious amount of possibilities of factors responsible for phishing . Rough sets can be a powerful tool ,when working on such kind of Applications containing vague or imprecise data. This paper proposes an approach towards Phishing Detection Using Rough Set Theory. The Thirteen basic factors, directly responsible towards Phishing, are grouped into four Strata. Reliability Factor is determined on the basis of the outcome of these strata , using Rough Set Theory . Reliability Factor determines the possibility of a suspected site to be Valid or Fake. Using Rough set Theory most and the least influential factors towards Phishing are also determined.**

## I. Introduction

Phishing is an attack method ,which involves tricking someone by sending spam mails or instant messages, leading him to fraudulent websites ,to acquire his confidential details like password, bank account number, credit card number, social security number etc. It is roughly estimated around 1.2 million U.S. computer users suffered losses due to phishing totalling to about US$929 million between May 2004 and May 2005. The U.S. Businesses lose around US$2 billion per year due to their clients becoming victims to Phishing [1]. Phishing has now emerged as an ubiquitous internet fraud.

Various techniques have been developed to tackle Phishing like informing users about Phishing , Anti-Phishing softwares , Augmented password logins etc. Currently there are many softwares to combat Phishing sites. Ten Anti-Phishing softwares were tested for their efficiency in detecting phishing , Only one out of these ten softwares could detect Phishing sites with more than 60% accuracy [2]. The reason for this inefficiency is due to the fact, that parameters involved in phishing detection are vague. Phishing detection is a complex and intricate process which involves many factors. Hence, we argue that there is need for a better approach towards Phishing detection.

In this paper we will discuss about the vagueness of the parameters involved in phishing. The paper proposes a rough set based algorithm to determine a reliability factor of a given website. Reliability factor indicates the authenticity of the website . The paper proposes Rough Set theory for better analysis of the vagueness in parameters involved in Phishing detection in real time and thereby producing more accurate results for Phishing detection.

## II. Various methods for phishing attacks

Phishing is a method to steal confidential information from user by deceptive means .Apart from hacking, Phishing is an emerging field in the cybercrime nowadays. As the technology is becoming more and more sophisticated, so are the techniques used in phishing . Some of the basic techniques used in phishing are listed below.

### A. Spam E-mails

This is one of the most commonly used method in phishing . Phishers sends spam emails to millions of user .This spam mail may mimic an authentic mail. They may even use the company logos that are identical to the legitimate ones. Most of these mail may link to a fraudulent page or may require user to input his or her personal information like user name , passwords , etc. Most of these spam mails may persuade unwary users to take immediate action.

### B. Instant Messaging

It is estimated that 95% of workers will use instant messaging as their primary interface for computer based real - time communications. In February 2009 many users of Gmail and Yahoo were targeted by many phishing attacks through instant messaging [4]. In this method of phishing User receives the messages with a link directing them to a fake phishing website which appears visually similar to the genuine website. User is then asked to fill his/her personal details. Users are vulnerable through this method as the links may have similar url , ip address etc.

## C. Content injection

In this method phishers may change a part of content of genuine page. The altered content in the page may lead user to a fraudulent site [3].

## D. Malware Phishing

In this method spoof mails are send to multiple users. These mails contains malwares attached to it. On clicking the links in the spoof mails the malware begins to function. Using these malwares phishers can gain sensitive information for their own purposes [3].

### III. ROUGH SET THEORY APPROACH

Rough set theory is a new mathematical approach towards imprecise or vague data. It was introduced by Zdzislaw Pawlak [5]. It provides an excellent alternative to the problems involving imprecise data, where conventional set theory fails. As mentioned above many anti-phishing approaches fail to provide the accurate results [2]. This happens because parameters involved in phishing detection are vague thereby producing inaccurate results. This Paper proposes a Rough set theory approach to obtain more accurate results for phishing detection.

Let U be a Universal set and R be an indiscernibility relation R such that $R \subseteq U \times U$ which represents lack of information about the elements of U. Let X be a subset of U.

Using Basic Rough Set theory concepts

*R-Lower approximation of X*

$R_*(x) = \bigcup_{x \in U} \{ R(x) : R(x) \subseteq X \}$

*R- Upper approximation of X*

$R^*(x) = \bigcup_{x \in U} \{ R(x) : R(x) \cap X \neq \phi \}$

*R-Boundary region of X*

$RN_R(X) = R^*(x) - R_*(x)$

If $RN_R(X) = \phi$ *(Set X is Crisp Set)*
If $RN_R(X) \neq \phi$ *(Set X is Rough Set)*

As the Data is imprecise as well as superfluous. Thus many of its redundancy can be removed through calculating Reduct of the sets.[6]
Also
Core $(X) = \bigcap$ Red $(X)$ where Core of X is set of all indispensible attributes of X. [5]
Information is represented in form of two attributes Condition and Decision attributes.
All the inputs or parameters comprises of Conditions whereas Output comprises of Decision.

A decision rule is determined by each row. A set of Decision rules is known as Decision Algorithm. After Determining the Reduced set the respective Decision Algorithm is calculated [6].

Positive Region of U/B with respect to A is the set of all elements of U that can be uniquely classified to blocks of partition U/D by means of A [5].

$POS_A(B) = \bigcup_{X \in U/I(D)} A_*(X)$

### IV. ROUGH SET THEORY APPLIED TO PHISHING DETECTION

Approach here is to apply Rough set algorithm for Detection of Phishing by determining a reliability factor. Reliability Factor is used to predict the risk of phishing. There are many factors which determines the authenticity of the websites. Here thirteen basic factors will be considered and these will be the parameters for evaluating the reliability factor. These thirteen parameters will be further categorized into four strata.

**Stratum A** :-
a. Using Long URL as a link.
b. Using IP address rather than DNS name.
c. Large Number of Dots in IP Address.
d. Using Modified Port Number.

**Stratum B**:-
a. Suspicious SSL Certificate.
b. Age of Domain is Less than 6 months.
c. Unsecured Page.

**Stratum C**:-

a. Taking Longer time to access accounts.
b. Using Java Scripts to hide information.
c. Using Pop-Up Windows.

**Stratum D**:-
a. Visual similarity to other pages.
b. URL present in Google's Blacklist.
c. Redirected Pages.

These four Strata will be analysed independently for Phishing and then the results of these four strata will be combined to evaluate the reliability factor using Rough Set.

## A. Analyzing Stratum A

The Table A consists of four parameters. These parameters can have two input values yes or no. Depending upon these values on the parameters, the output known as the phishy status is generated.

TABLE I

| Using Long URL as a link | Using IP address rather than DNS | Large Number of dots in IP Address | Using modified port number | Phishy status of Stratum A |
|---|---|---|---|---|
| No | No | No | No | Valid |
| No | No | No | Yes | Valid |
| No | No | Yes | No | Valid |
| No | No | Yes | Yes | Suspicious |
| No | Yes | No | No | Valid |
| No | Yes | No | Yes | Suspicious |
| No | Yes | Yes | No | Suspicious |
| No | Yes | Yes | Yes | Valid |
| Yes | No | No | No | Valid |
| Yes | No | No | Yes | Suspicious |
| Yes | No | Yes | No | Suspicious |
| Yes | No | Yes | Yes | Fake |
| Yes | Yes | No | No | Suspicious |
| Yes | Yes | No | Yes | Suspicious |
| Yes | Yes | Yes | No | Fake |
| Yes | Yes | Yes | Yes | Fake |

*B. Analyzing Stratum B*

Similarly Stratum B is divided into three parameters. These parameters have two input values yes or no. Depending upon these values on the parameters the Phishy state of Stratum B is generated.

TABLE II

| Suspicious SSL certificate | Age of Domain is less than 6 Months | Unsecured page | Phishy status of Stratum B |
|---|---|---|---|
| No | No | No | Valid |
| No | No | Yes | Suspicious |
| No | Yes | No | Suspicious |
| No | Yes | Yes | Fake |
| Yes | No | No | Fake |
| Yes | No | Yes | Fake |
| Yes | Yes | No | Fake |
| Yes | Yes | Yes | Fake |

*C. Analyzing Stratum C*

Similarly Stratum C is divided into three parameters. These parameters have two input values yes or no. Depending upon these values on the parameters the Phishy state of Stratum C is generated.

TABLE III

| Taking Longer Time to access Accounts | Using Java scripts to hide information | Using Pop-Up Windows | Phishy Status of C |
|---|---|---|---|
| No | No | No | Valid |
| No | No | Yes | Suspicious |
| No | Yes | No | Suspicious |
| No | Yes | Yes | Fake |
| Yes | No | No | Fake |
| Yes | No | Yes | Fake |
| Yes | Yes | No | Fake |
| Yes | Yes | Yes | Fake |

*D. Analyzing Stratum D*

Similarly Stratum D is divided into three parameters. These parameters have two input values yes or no. Depending upon these values on the parameters the Phishy state of Stratum D is generated.

TABLE IV

| Visual Similarity | URL is present in Google's Blacklist | Redirected pages | Phishy status of Stratum D |
|---|---|---|---|
| No | No | No | Valid |
| No | No | Yes | Suspicious |
| No | Yes | No | Suspicious |
| No | Yes | Yes | Fake |
| Yes | No | No | Fake |
| Yes | No | Yes | Fake |
| Yes | Yes | No | Fake |
| Yes | Yes | Yes | Fake |

*E. Determining the Reliability Factor*

The reliability factor will be determined by taking the inputs of all four Strata, each inputs of these four Strata is combined and Rough Set Theory is applied to obtain a corresponding reliability factor. Reliability factor determines the probability of a suspected site to be Reliable or Unreliable.

The Table below shows the Reliability Factor on the basis of Phishy Status of four Strata.

TABLE V

| Phishy State of Stratum A | Phishy State of Stratum B | Phishy State of Stratum C | Phishy State of Stratum D | Reliabilty Factor |
|---|---|---|---|---|
| Valid | Valid | Valid | Valid | Reliable |
| Valid | Valid | Valid | Suspicious | Reliable |
| Valid | Valid | Suspicious | Fake | Unreliable |
| Valid | Valid | Suspicious | Valid | Reliable |
| Valid | Suspicious | Fake | Suspicious | Reliable |
| Valid | Suspicious | Fake | Fake | Unreliabl |

| | s | | | e | |
|---|---|---|---|---|---|
| Valid | Suspicious | Valid | Valid | Reliable |
| Valid | Suspicious | Valid | Suspicious | Unreliable |
| Valid | Fake | Suspicious | Fake | Unreliable |
| Suspicious | Fake | Suspicious | Valid | Unreliable |
| Suspicious | Fake | Fake | Suspicious | Unreliable |
| Suspicious | Fake | Fake | Fake | Unreliable |
| Suspicious | Valid | Valid | Valid | Reliable |
| Suspicious | Valid | Valid | Suspicious | Reliable |
| Suspicious | Valid | Suspicious | Fake | Unreliable |
| Suspicious | Valid | Suspicious | Valid | Reliable |
| Suspicious | Suspicious | Fake | Suspicious | Unreliable |
| Suspicious | Suspicious | Fake | Fake | Unreliable |
| Fake | Suspicious | Valid | Valid | Unreliable |
| Fake | Suspicious | Valid | Suspicious | Unreliable |
| Fake | Fake | Suspicious | Fake | Unreliable |
| Fake | Fake | Suspicious | Valid | Unreliable |
| Fake | Fake | Fake | Suspicious | Unreliable |
| Fake | Fake | Fake | Fake | Unreliable |
| Valid | Valid | Valid | Suspicious | Unreliable |
| Suspicious | Valid | Valid | Suspicious | Unreliable |

### F. Decision Algorithm

Stratum A , Stratum B , Stratum C and Stratum D are condition attributes . R Factor is a Decision attribute.

*If ("Stratum D"=Fake) then ("R Factor"=Unreliable)*

*If ("Stratum B"=Fake) then ("R Factor"=Unreliable)*

*If ("Stratum A"=Fake) then ("R Factor"=Unreliable)*

*If ("Stratum B"=Valid) & ("Stratum D"=Valid) then ("R Factor"=Reliable)*

*If ("Stratum B"=Valid) & ("Stratum D"=Suspicious) then ("R Factor"=Unreliable)*

*If ("Stratum A"=Suspicious) & ("Stratum C"=Fake) then ("R Factor"=Unreliable)*

*If ("Stratum A"=Valid) & ("Stratum D"=Valid) then ("R Factor"=Reliable)*

*If ("Stratum B"=Suspicious) & ("Stratum C"=Valid) & ("Stratum D"=Suspicious) then ("R Factor"=Unreliable)*

*If ("Stratum A"=Suspicious) & ("Stratum C"=Valid) & ("Stratum D"=Suspicious) then ("R Factor"=Unreliable)*

*If ("Stratum A"=Suspicious) & ("Stratum B"=Suspicious) then ("R Factor"=Unreliable)*

*If ("Stratum A"=Valid) & ("Stratum C"=Fake) & ("Stratum D"=Suspicious) then ("R Factor"=Reliable)*

*If ("Stratum A"=Suspicious) & ("Stratum C"=Valid) & ("Stratum D"=Valid) then ("R Factor"=Reliable)*

### G. Reduct Set

| Reduct | Positive region |
|---|---|
| "Stratum A", "Stratum D" | 0.5769230769230769 |
| "Stratum A", "Stratum C", "Stratum D" | 0.7307692307692307 |
| "Stratum A", "Stratum C" | 0.38461538461538464 |
| "Stratum A", "Stratum B" | 0.46153846153846156 |
| "Stratum A" | 0.23076923076923078 |
| "Stratum B", "Stratum D" | 0.6153846153846154 |
| "Stratum B", "Stratum C", "Stratum D" | 0.6923076923076923 |
| "Stratum B" | 0.3076923076923077 |
| "Stratum D" | 0.3076923076923077 |

The Positive region determines the probability of a suspected site to be Phishy on the basis of decision factors which also determined Reliability Factor.

Analysing the above Reduct set we may infer the following possible outcomes -

A. Stratum A is the least influencing factor for detecting a site to be fraudulent i.e. factors such as using Long URL as a link , using IP address rather than DNS name, large Number of Dots in IP Address and using Modified Port Number should be given least weightage when inspecting a suspected Phishy site.

B. Stratum A , Stratum C and Stratum D combined are the most influencing factor for detecting a site to be fraudulent.

If all these three Strata confirm a site to be non valid then the site is most likely to be a Phishy site.

## V. CONCLUSIONS

In this paper we implemented Rough set theory in Phishing Detection Algorithm and got more accurate results .Using Rough Set Theory the huge data was reduced in a more organized and systematic data. Hence ,we determined a Reliability factor using Rough Sets . This Reliability Factor helps us in determining whether the suspected site is Valid or Fake. Using Rough Set we also elucidated ,the Most and the Least influencing factors while detecting a Phishy site. The Limitations of this approach is that,, it only determines the probability of a site to be reliable or unreliable. An site which is determined to be unreliable does not necessarily means that the site might actually be Unreliable. It only has a higher probability to be Unreliable or more likely to be Unreliable.

As the technology is burgeoning at a tremendous rate, so many new factors may emerge which may influence Phishing in one way or the other . Thus this approach does not guarantee success every time but , this approach enunciates an alternate and possibly a more efficient method in detecting Phishing by using Rough Sets.